\documentclass[10pt,twocolumn,letterpaper]{article}

\usepackage[pagenumbers]{iccv} 

\usepackage{array}
\usepackage{multirow}
\usepackage{graphicx}
\usepackage{tikz}
\usepackage{lipsum}  
\usepackage{svg}

\usepackage{svg}

\definecolor{iccvblue}{rgb}{0.21,0.49,0.74}
\usepackage[pagebackref,breaklinks,colorlinks,allcolors=iccvblue]{hyperref}

\def\paperID{14488} 
\def\confName{ICCV}
\def\confYear{2025}

\title{VideoMultiAgents: A Multi-Agent Framework for Video Question Answering}

\author{
Noriyuki Kugo$^1$, 
Xiang Li$^2$, 
Zixin Li$^2$, 
Ashish Gupta$^1$, 
Arpandeep Khatua$^2$,\\
Nidhish Jain$^2$, 
Chaitanya Patel$^2$, 
Yuta Kyuragi$^{2,3}$,
Yasunori Ishii$^4$, \\
Masamoto Tanabiki$^1$,
Kazuki Kozuka$^4$, 
Ehsan Adeli$^2$ \\
\smallskip
\small $^1$Panasonic Connect Co., Ltd.,
\small $^2$Stanford University, \\
\small $^3$Panasonic R\&D Company of America, 
\small $^4$Panasonic Holdings Corporation
}

\begin{document}
\maketitle
\begin{abstract}

Video Question Answering (VQA) inherently relies on multimodal reasoning, integrating visual, temporal, and linguistic cues to achieve a deeper understanding of video content. However, many existing methods rely on feeding frame-level captions into a single model, making it difficult to adequately capture temporal and interactive contexts. To address this limitation, we introduce VideoMultiAgents, a framework that integrates specialized agents for vision, scene graph analysis, and text processing. It enhances video understanding leveraging complementary multimodal reasoning from independently operating agents.
Our approach is also supplemented with a question-guided caption generation, which produces captions that highlight objects, actions, and temporal transitions directly relevant to a given query, thus improving the answer accuracy. Experimental results demonstrate that our method achieves state-of-the-art performance on Intent-QA (79.0\%, +6.2\% over previous SOTA), EgoSchema subset (75.4\%, +3.4\%), and NExT-QA (79.6\%, +0.4\%).
The source code is available at \url{https://github.com/PanasonicConnect/VideoMultiAgents}.

\end{abstract}

\begin{figure}[t]
  \centering
  \includegraphics[width=1\linewidth]{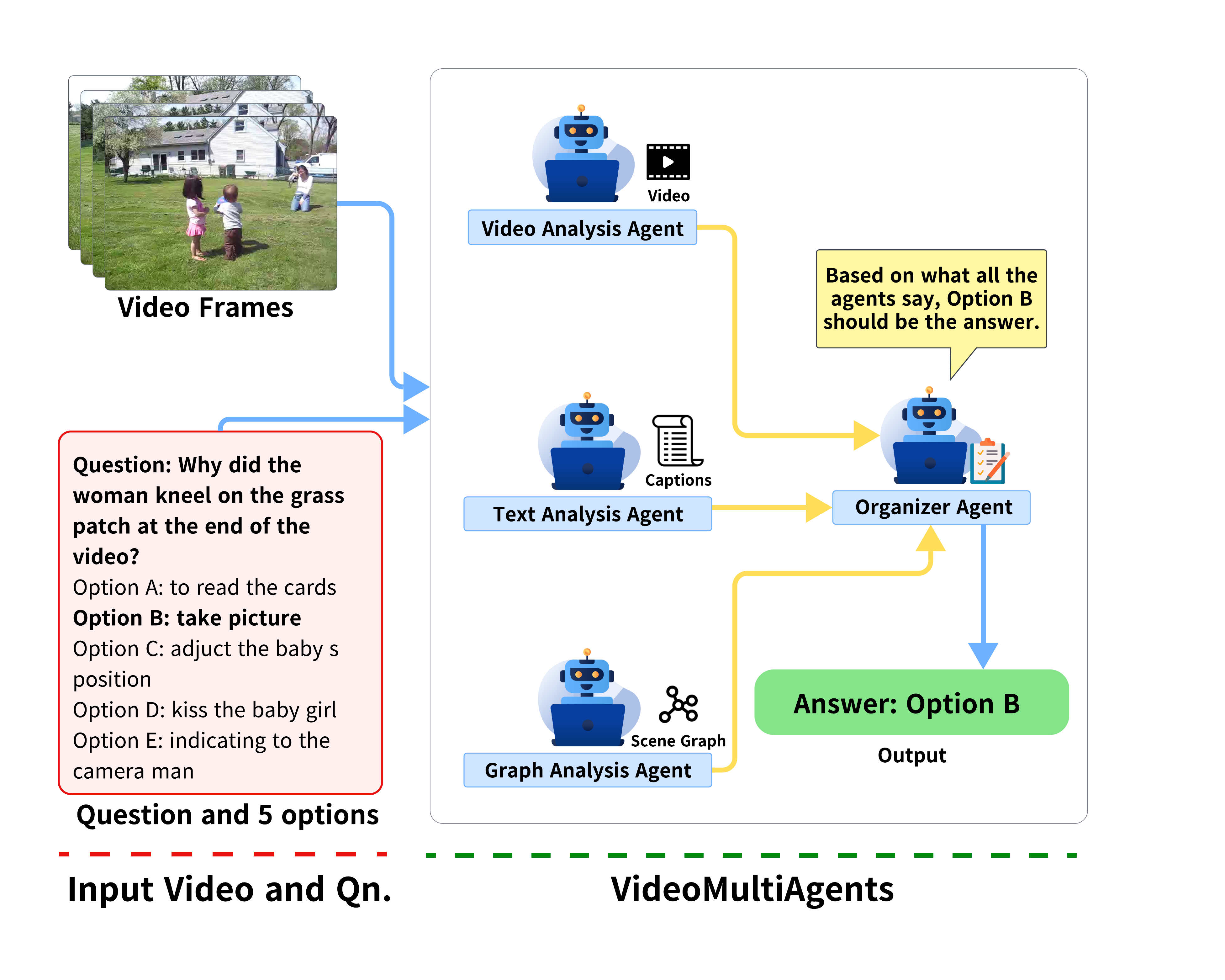} 
  \caption{Overview of VideoMultiAgents: Integrating Video, Captions, and Scene Graphs through Multi-Agent Collaboration for accurate Video Question Answering. Blue arrows indicate inputs and outputs to and from the VideoMultiAgents, while yellow arrows represent interactions and information exchanges between individual agents, coordinated by the Organizer Agent to determine the final answer.}
  \label{fig:masteaser}
\end{figure}

\section{Introduction}
\label{sec:intro}
Video understanding is a challenging task that requires integrating visual and textual information while reasoning about temporal sequences to answer queries in natural language. Recent advances in large language models (LLMs) have dramatically enhanced their ability to process multimodal data. These models now seamlessly integrate visual and textual information, allowing them to handle complex queries that require understanding both contexts simultaneously~\cite{li2023blip, wang2023videollava}. However, existing video question answering (VQA) approaches still face critical limitations. Most notably, current methods typically rely on captioning systems to convert long videos into text for downstream reasoning~\cite{Wang2024VideoTreeAT, wang2024videoagent, Zhang2023ASL}, yet these systems struggle to accurately capture temporal dynamics, causal relationships, and detailed visual context. This is largely due to the inherent loss of important visual information when compressing video frames into dense captions and their dependence on general-purpose captions that aren't tailored to specific questions.

Numerous approaches have been proposed to overcome these issues. LLoVi~\cite{Zhang2023ASL} introduced a text-centric approach, converting video frames into captions and processing them through an LLM, which efficiently reduces VQA tasks to language-based reasoning. However, because captions are generated independently of the specific question, critical details required to answer complex questions might be missed. VideoAgent~\cite{wang2024videoagent} introduced an innovative agent-based iterative frame selection method driven by an LLM, significantly reducing computational costs and improving frame efficiency. Despite this advancement, VideoAgent relies solely on textual captions as its primary modality, potentially missing essential visual and spatial details critical for precise reasoning. VideoTree~\cite{Wang2024VideoTreeAT} employs a hierarchical structure for efficient context retention, but similarly generates general, question-independent captions, limiting its ability to tailor information to specific questions. VDMA~\cite{kugo2024vdma} improves performance by dynamically generating specialized agents for each query, but its effectiveness strongly depends on selecting suitable agent roles, potentially limiting the full utilization of their modality-specific strengths.


To overcome these challenges, we propose \textbf{VideoMultiAgents}, a novel multi-agent framework for video question answering (see Fig.~\ref{fig:masteaser}). In this framework, an \textit{organizer agent} serves as a central integrator, synthesizing information from multiple independent modality-specific agents, each specializing in different aspects of multimodal reasoning (e.g., visual, textual, or graph analysis).
Each modality-specific agent independently conducts inference based on its assigned modality and reports its findings to the organizer. The organizer then aggregates these outputs, resolves potential inconsistencies, and determines the final answer. This structured approach substantially enhances multimodal reasoning by leveraging the complementary strengths of specialized agents while maintaining coherence in the final decision-making process.

Our main contributions are summarized as follows.

\begin{itemize}
\item \textbf{Modality-Specific Agents with Organizer:}
Our framework consists of three specialized agents: (1) a Text Analysis Agent that generates and analyzes question-guided textual captions, (2) a Video Analysis Agent that extracts detailed visual features from video frames, and (3) a Graph Analysis Agent that constructs structured scene graphs that capture temporal and spatial relationships. Additionally, an Organizer Agent collects and consolidates the outputs from these agents, ensuring consistency and resolving any contradictions before deriving the final answer.

\item \textbf{Question-Guided Caption Generation:}
Unlike previous methods, our caption generation explicitly leverages key nouns and verbs derived from the query to create captions highly relevant to the question. This approach substantially reduces irrelevant information and emphasizes essential entities and actions, directly addressing the shortcomings of generic caption-based methods.

\item \textbf{Scene graphs for enhancing video understanding:}
Capturing the interactions among people and objects, as well as the depicted scenes and their temporal evolution, is essential for robust video recognition. Scene graphs, which encode people and objects as nodes and their relationships and actions as edges, provide a concise yet explicit representation of a video’s content. By systematically organizing complex spatiotemporal information, these graphs not only facilitate the identification of causal relationships and temporal dynamics, but also empower models to interpret and describe video content more comprehensively.


\end{itemize}
Through comprehensive evaluations of VQA benchmark datasets, VideoMultiAgents demonstrates superior zero-shot performance compared to existing state-of-the-art methods, achieving 79.0\% (+6.2\% improvement over previous SOTA) in Intent-QA, 75.4\% in EgoSchema subset (+3.4\%), and 79.6\% in NExT-QA (+0.4\%). These results clearly highlight our framework's robust generalization capability and adaptability to novel tasks without task-specific training, showcasing its practical potential for diverse real-world scenarios.
Importantly, our multi-agent framework effectively combines multiple modalities, significantly enhancing reasoning performance and adaptability. This approach is inherently scalable, enabling easy integration of additional modality-specific agents for new tasks or emerging challenges.


\section{Related Work}
\label{sec:related_work}

This section categorizes prior research on Video Question Answering into four key perspectives, each explained in the following subsections. 

\subsection{Classification of VQA approaches}

VQA can be broadly grouped into several paradigms, including end-to-end~\cite{Wang2022InternVideoGV, Ye2023mPLUGOwlME,Lin2023VideoLLaVALU,Zhang2023VideoLLaMAAI, Song2023MovieChatFD, Maaz2023VideoChatGPTTD}, caption-based~\cite{Wang2024VideoTreeAT, Zhang2023ASL,Wang2023LifelongMemoryLL,zhang2024hcqaego4degoschema}, graph-based~\cite{cherian2021adaptive,rodin2023actionscenegraphslongform,10203761}, 
multimodal fusion~\cite{Wang2022InternVideoGV, Fan2024VideoAgentAM, Lin2023MMVIDAV}, 
and agent-based~\cite{wang2024videoagent,Fan2024VideoAgentAM,kugo2024vdma} methods. Of these, VLM-based, caption-based, and agent-based approaches have shown particular promise in improving the performance, adaptability, and interpretability of VQA systems.
In the following sections, we focus on vision-language models, caption-based strategies, and multi-agent reasoning, as these have shown promising potential for improving VQA performance.

\subsection{Vision-Language Model-based methods}

Vision-Language Models (VLMs) integrate visual and textual modalities into a joint embedding space, achieving strong performance in various multimodal tasks. Recent VLMs have enhanced video understanding capabilities through advanced representation learning and instruction-tuning approaches.
InternVideo \cite{Wang2022InternVideoGV} learns generalized video-language representations using masked video modeling and contrastive learning, enabling strong zero-shot performance across tasks. However, it relies heavily on extensive pre-training on large-scale multimodal datasets, limiting its scalability and adaptability when handling diverse downstream VQA tasks.
mPLUG-Owl \cite{Ye2023mPLUGOwlME} addresses multimodal integration using a two-stage training (pre-training and instruction tuning), effectively enhancing adaptability to complex multimodal tasks. However, this approach primarily handles static images, which limits its ability to manage the temporal dynamics inherent in videos.
Video-LLaVA \cite{Lin2023VideoLLaVALU} extends multimodal capabilities by jointly training on image and video data, thereby improving temporal reasoning. While effective, it requires extensive multimodal instruction-following data for fine-tuning, posing scalability concerns and challenges in generalizing quickly to novel tasks.
Video-LLaMA \cite{Zhang2023VideoLLaMAAI} introduces a multimodal LLM that integrates video and audio encoders. Although powerful in audio-visual integration and temporal reasoning, its reliance on specific pre-trained visual and audio encoders limits flexibility in adapting to novel modalities or domains, impacting its adaptability.

In general, existing VLMs significantly advance multimodal understanding, but encounter key limitations, including restricted modality support, limited adaptability to specific task instructions without substantial retraining, and inherent scalability challenges due to extensive training requirements.
In contrast, our proposed VideoMultiAgents framework explicitly addresses these limitations by leveraging specialized agents that adaptively interact, dynamically integrating multimodal data to effectively handle diverse and complex VQA tasks.

\subsection{Caption-based methods.}

Caption-based methods tackle video question answering (VQA) by first converting visual content from videos into textual descriptions (captions), and then leveraging LLMs or other NLP models to answer queries. By translating the VQA task into a purely textual question-answering problem, these methods directly exploit the remarkable reasoning capabilities of modern LLMs, thus achieving significant advances in performance.

VideoTree \cite{Wang2024VideoTreeAT} addresses the challenges associated with lengthy videos by clustering similar frames and extracting only the clusters that are relevant to the question, thus improving both efficiency and accuracy. LLoVi \cite{Zhang2023ASL} generates frame-level captions independently using an image captioning model and subsequently uses an LLM to derive answers. LifelongMemory \cite{Wang2023LifelongMemoryLL} further refines this paradigm by filtering out redundant or noisy captions to ensure that the extracted information is concise and relevant. HCQA \cite{zhang2024hcqaego4degoschema} integrates few-shot prompting strategies to enhance caption quality and contextual relevance, resulting in notable performance improvements.

However, caption-based methods commonly face two critical limitations. First, captions are typically generated without explicit consideration of the query context, leading to the potential omission of question-specific visual details critical for accurate answers. Second, the frame-by-frame captioning approach may not adequately represent temporal relationships and dynamics in videos, limiting the model’s ability to handle queries that explicitly depend on temporal understanding. Addressing these shortcomings remains essential for advancing caption-based VQA methodologies.

\subsection{Agent-based methods.}

Moreover, agent-based methods have recently emerged in the video VQA domain. For instance, VideoAgent~\cite{wang2024videoagent} leverages LLMs to efficiently select relevant frames based on the given query, significantly improving computational efficiency. However, since it relies exclusively on textual information after frame selection, it may miss essential visual and spatial details crucial for accurate answers. Similarly, VideoAgent~\cite{Fan2024VideoAgentAM} employs a single agent driven by LLM capable of using multiple external tools, but its single agent structure limits scalability and flexibility as the number of tools increases. In contrast, VDMA~\cite{kugo2024vdma} dynamically generates specialized agents for different subtasks per query, enhancing performance but highly depending on agent role selection, potentially underutilizing modality-specific strengths.

In response to these challenges, our proposed \textbf{VideoMultiAgents} framework introduces multiple agents specialized by modality—textual, visual, and graph-based—and coordinates them through an Organizer. This design effectively exploits modality-specific strengths, improving both efficiency and adaptability in addressing various video question-answer tasks.

\begin{figure*}
  \centering
  \includegraphics[width=\textwidth]{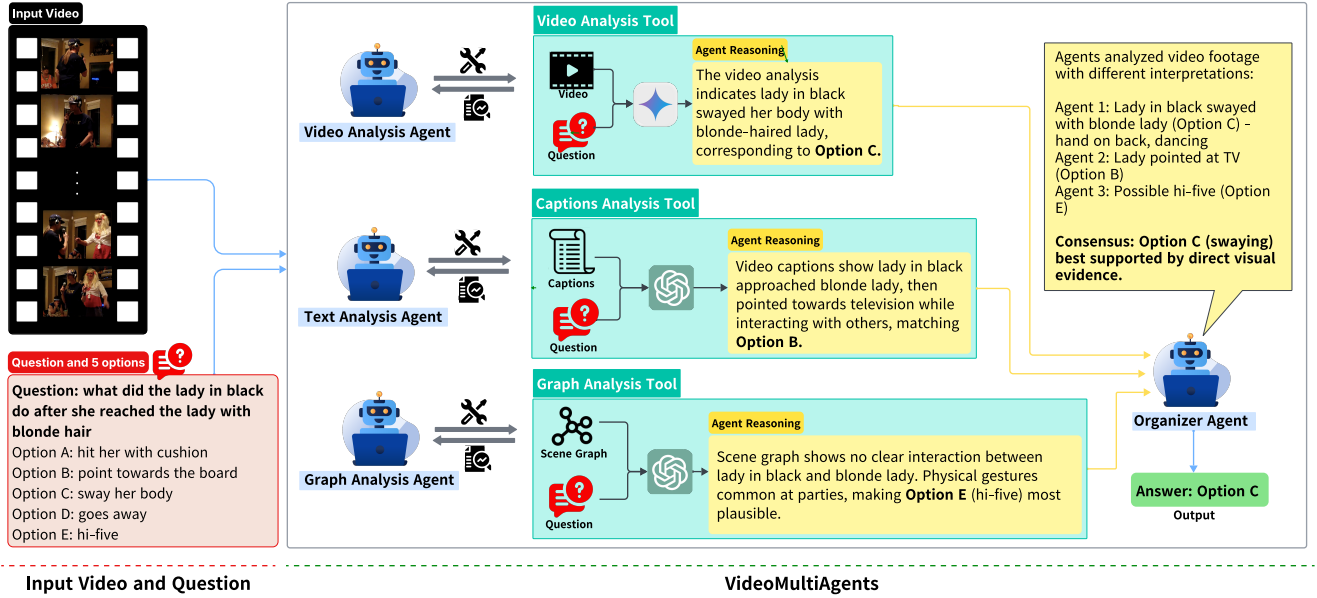}
  \caption{Detailed Architecture of a VideoMultiAgents for Video Question Answering: Each specialized agent independently analyzes its assigned modality—text, video, or scene graph—based on the input video and question, generating its own response. The Text Analysis Agent utilizes question-guided captions to extract key entities and actions. The Video Analysis Agent identifies objects, interactions, and temporal events. The Graph Analysis Agent constructs structured representations of object relationships and causal dynamics. The Organizer Agent integrates these independent responses to synthesize the final answer.}
  \label{fig:mas_detailed}
\end{figure*}

\section{Method}
\label{sec:method}

In this section, we introduce VideoMultiAgents, a multi-agent framework for Video Question Answering (VQA), illustrated in Figure \ref{fig:mas_detailed}. Our approach features a Organizer Agent that integrates the outputs from three specialized agents—Text Analysis Agent, Video Analysis Agent, and Graph Analysis Agent—to generate a coherent final answer. 

\subsection{MultiAgent Framework}

Before finalizing our current framework design, we explored several alternative multi-agent architectures, including star structures (allowing dynamic communication between agents and the organizer) and debate structures (enabling sequential critique among agents). However, allowing the organizer to access answers too early or enabling direct communication between agents led to biased answer generation and error propagation, ultimately degrading individual agent performance and, consequently, the overall system's effectiveness. Instead, we found that a report structure—where the Organizer Agent collects insights from independent, modality-specialized agents—leads to a more robust and balanced decision-making process. Each agent operates independently within its modality, preventing interference while maintaining specialization. The Organizer Agent then aggregates the results, resolves conflicts, and determines the final answer. This approach ensures a stable reasoning process, avoids single-agent domination, and fully leverages each agent’s strengths. We further analyze this trend in our ablation studies~\ref{subsec:Effectiveness of Multi-Agent Architectures}, where we compare the impact of various communication structures on system effectiveness.

As shown in Figure \ref{fig:mas_detailed}, our system processes an input video and a question with multiple-choice answers. Each specialized agent analyzes its assigned aspect of the video within its modality, returning partial answers and supporting evidence. The Organizer Agent then aggregates and reconsiders the responses from all agents. By evaluating the reasoning from each agent, it synthesizes the most accurate final answer and selects the best matching multiple-choice option.


\subsection{Organizer Agent}
The Organizer Agent is responsible for integrating the outputs from multiple specialized agents to generate a coherent and well-reasoned final answer.
Each specialized agent runs independently, processing its assigned modality—text, video, or graph—and returning a structured response. The Organizer Agent then evaluates these responses, ensuring consistency, resolving conflicts, and synthesizing the final answer.

To prevent interference between agents, communication is restricted to interactions between each agent and the Organizer. This design ensures that each agent keeps its own independent memory while diving deeply into its own modality. By integrating independently derived insights from specialized agents, the framework effectively handles a wide variety of video-based questions.

\subsection{Specialized Agents}
Each specialized agent is tasked with interpreting a specific modality using a modality-specific analysis tool. The agent can invoke the tool repeatedly with different arguments to look for evidence in the video and make a final decision when it is confident in the answer.
\begin{itemize}
    \item \textbf{Text Analysis Agent:}
    
    This agent is responsible for using video captions to solve the question. It has access to a caption analysis tool which generates question-guided captions for each segment of the video.
    
    \item \textbf{Video Analysis Agent:} 
    
    This agent is responsible for interpreting the raw visual content of the video and extracting the key evidence needed to answer the question. With access to a video analysis tool powered by vision language models (VLM), the agent can use the tool to learn about various aspects of the video.
    \item \textbf{Graph Analysis Agent:} 

    This agent is designed to construct structured representations of objects, actions, and their relationships over time, improving the ability to reasoning about causal relationships, spatial dependencies, and temporal dynamics within the video, inspired by recent advancements in fine-grained video question answering MOMA-QA~\cite{dai2023momaqa} and EASGs~\cite{rodin2023actionscenegraphslongform}. The video is processed into small temporal segments and for each segment, a scene graph is generated. A scene graph is a structured representation where nodes correspond to objects, characters, or key entities, and edges represent their relationships and actions. Unlike video analysis agent, which primarily focuses on pixel-level features, this agent abstracts interactions into a graph-based structure, capturing how entities interact across different moments in time. 
    
\end{itemize}

\subsection{Question-Guided Caption Generation}

In this section, we introduce our second contribution: Question Guided Captioning. Recently, image captioning techniques, which convert each video frame into textual information, have been widely adopted as encoders in video recognition tasks. For example, LloVi~\cite{Zhang2023ASL} converts images into text and then feeds those captions into a large language model (LLM), achieving strong performance through a relatively simple pipeline.
However, one major concern with such approaches is that, if the generated captions fail to emphasize the crucial elements required to answer a question, subsequent reasoning tasks like VQA can be adversely affected. In contrast, when humans watch a video with the aim of answering questions about its content, they typically pay attention to objects and actions relevant to the question.

To address this issue, our approach first analyzes the question and its answer choices to extract important nouns, verbs, and other keywords. We then guide a VLM to generate captions that highlight these extracted terms, thereby producing text closely aligned with the question.
In addition, instead of generating captions frame by frame, we process several frames at once to capture temporal information. This strategy is particularly effective for understanding how actions evolve or how objects move over time, which is essential for accurately identifying and interpreting verbs.
By focusing on question-related keywords and incorporating time-based context, our question-guided captioning method yields captions that filter out irrelevant details and emphasize the core elements needed for VQA. As a result, we expect more accurate performance in downstream tasks compared to conventional, general-purpose captioning methods.

\section{Experiments}
\label{sec:experiments}
In this section, we present the evaluation of our proposed VideoMultiAgents in several VQA datasets. 
We first describe the datasets used for our experiments in Section~\ref{subsec:datasets}. Next, we provide implementation details on an instantiation of our proposed VideoMultiAgents architecture in Section~\ref{subsec:implementation}.
Then, we compare our approach with existing state-of-the-art methods in Section~\ref{sec:results}, followed by ablation studies to analyze the contribution of each component in Section~\ref{sec:ablation}. 
Finally, we provide further discussion and illustrative case studies in Section~\ref{sec:casestudy}.

\subsection{Datasets}
\label{subsec:datasets}

We evaluated VideoMultiAgents on three video quality assurance datasets: 
NExT-QA~\cite{xiao2021next}, Intent-QA~\cite{10376648}, and EgoSchema~\cite{mangalam2023egoschema}. These datasets vary in video perspective, duration, and reasoning complexity, offering a diverse benchmark for assessing video-language models.

\noindent\textbf{NExT-QA}~\cite{xiao2021next} provides roughly 52k question-answer pairs in 5,440 short third-person videos.
We focus on the validation split of 570 videos and 4,996 questions.
This dataset emphasizes causal and temporal queries (e.g., ``Why did X happen?'' or ``What happened after Y?''), thus requiring models to detect event progressions and motivations in typical real-world settings.

\noindent\textbf{Intent-QA}~\cite{10376648} is built on top of NExT-QA videos focusing on inference type questions. Intent-QA comprises of Causal(why, how) and Temporal(before, after) categories and  contains 567 videos and 2,134 multiple choice questions in the test split. 

\noindent\textbf{EgoSchema}~\cite{mangalam2023egoschema} is an egocentric benchmark derived from Ego4D, featuring first-person videos that each lasting about three minutes.
It contains 5,031 multiple choice QA pairs (\emph{fullset}), with an officially released \emph{subset} of 500 questions for public evaluation.
Because the clips are long and captured from a personal viewpoint, the models must demonstrate robust long-horizon temporal reasoning and viewpoint adaptation to integrate events scattered throughout the 3-minute videos.


\subsection{Implementation Details}
\label{subsec:implementation}

We implemented our proposed VideoMultiAgents framework using the open-source Lang Graph library, which provides a graph-based infrastructure for orchestrating multiple tools and agents in conjunction with an LLM. For all agents and the text and graph analyzer tools, we use OpenAI's GPT-4o~\cite{OpenAI2024GPT4o} (version gpt-4o-2024-08-06) as the LLM. These agents internally call dedicated tools to generate captions and scene graphs. When producing Question-Guided Captions, we sample the input video at 1 fps and generate captions in 5-frame segments with a 1-frame overlap.
To handle scene graph generation, we sample sequential chunks of video frames that share similar caption content. When creating each new scene graph, the system references the previous chunk’s scene graph if one exists and the captions to help generate relationship triplets. This approach allows the system to maintain contextual continuity while building a structured representation of the video content as it progresses through the footage. To handle video analysis, our video agent leverages Google’s Gemini 2.0-flash, which can directly process the raw video files along with queries generated by the agents.




\section{Results}
\label{sec:results}
We evaluated our proposed VideoMultiAgents framework on three widely-used benchmark datasets: NExT-QA, Intent-QA and EgoSchema.

First, Table~\ref{tab:nextqa} shows the zero-shot performance comparison on the NExT-QA dataset, which focuses on causal and temporal reasoning in daily activities recorded from a third-person perspective. Our method delivers robust performance across all categories, notably achieving accuracies of 80.5\%, 75.7\%, and 84.7\% in the Causal, Temp, and Descriptive categories, respectively, and an overall accuracy of 79.6\%. \textbf{Our approach achieves state-of-the-art performance on NExT-QA}, improving the overall accuracy from Tarsier~\cite{wang2024tarsierrecipestrainingevaluating}’s 79.2\% by 0.4\%. In particular, we do not rely on additional training data, whereas Tarsier utilized additional data, thus establishing a new benchmark under more constrained training conditions.

Next, Table~\ref{tab:intentqa} shows the zero-shot performance comparison on the Intent-QA dataset, which primarily emphasizes inference-based reasoning across three question categories—CW (Causal-Why), CH (Causal-How), and TP\&TN (Temporal). Our framework achieves an overall accuracy of 79.0\%, representing a 6.2\% gain over the previous state-of-the-art method, VideoINSTA~\cite{Liao2024VideoINSTAZL} (72.8\%). \textbf{Our approach achieves state-of-the-art performance on Intent-QA}, underscoring the effectiveness of our multi-agent framework in addressing inference-driven queries.

Finally, Table~\ref{tab:egoschema} summarizes the results on the EgoSchema dataset, which features long-form egocentric videos and thus demands robust long-range temporal reasoning capabilities. Our VideoMultiAgents method achieves accuracies of 75.4\% on the subset and 68.0\% on the fullset, obtaining a \textbf{state-of-the-art performance with a +3.4\% improvement over previous methods in the EgoSchema subset.} This demonstrates that our approach achieves state-of-the-art performance on the EgoSchema subset, highlighting its effectiveness in long-range reasoning tasks.
While our method excels in the subset, it falls short in the full set compared to top-performing approaches such as HCQA~\cite{zhang2024hcqaego4degoschema} (75.0\%). Interestingly, HCQA exhibits a substantial gap between its subset score (58.8\%) and full set score (75.0\%), a trend also observed in other methods. This suggests that the distribution of question categories differs between the subset and the full set in EgoSchema, potentially influencing model performance. A more detailed analysis of these distributional differences could provide valuable insights for further improving the adaptability and robustness of our model across diverse question types.


Overall, our VideoMultiAgents framework demonstrated robust performance, particularly on questions requiring causal and inference-based reasoning, clearly validating the efficacy of integrating multimodal information through multiple specialized agents. Furthermore, our framework exhibited strong generalization capabilities in zero-shot scenarios, demonstrating promising applicability across diverse question types and video contexts.



\begin{table}
  \caption{Zero-Shot Performance on NExT-QA: A Comparison with State-of-the-Art Approaches\\
    {\footnotesize * Tarsier's results are reported with additional training data.}
  }
  \centering
  \setlength{\tabcolsep}{1pt}
  \begin{tabular}{lc|c|c|cc}
    \toprule
    \multirow{2}{*}{\textbf{Method}} & \multicolumn{4}{c}{\textbf{Accuracy (\%)}} \\ 
    \cmidrule(lr){2-5}
    & \textit{Caus.} & \textit{Temp.} & \textit{Desc.} & \textit{All} \\
    \midrule
    IG-VLM~\cite{kim2024image} & 72.2 & 65.7 & 77.3 & 70.9  \\ 
    VideoAgent~\cite{Wang2024VideoAgentLV} & 72.7 & 64.5 & 81.1 & 71.3  \\ 
    LVNet~\cite{Park2024TooMF} & 75.0 & 65.5 & 81.5 & 72.9  \\ 
    VideoTree~\cite{Wang2024VideoTreeAT} & 75.2 & 67.0 & 81.3 & 73.5  \\ 
    TS-LLAVA~\cite{qu2024tsllavaconstructingvisualtokens} & 74.6 & 68.2 & 81.5 & 73.6  \\ 
    LLoVi~\cite{Zhang2023ASL} & 73.7 & 70.2 & 81.9 & 73.8  \\ 
    ENTER~\cite{DBLP:journals/corr/abs-2501-14194} & 77.9 & 68.2 & 79.2 & 75.1  \\
    Tarsier*~\cite{wang2024tarsierrecipestrainingevaluating} & - & - & - & 79.2  \\
    \midrule
    \textbf{VideoMultiAgents (Ours)} & \textbf{80.5} & \textbf{75.7} & \textbf{84.7} & \textbf{79.6} \\
    \bottomrule
  \end{tabular}
  \label{tab:nextqa}
\end{table}


\begin{table}
  \caption{Zero-Shot Performance on Intent-QA: A Comparison with State-of-the-Art Approaches.} 
  \centering
  \setlength{\tabcolsep}{1pt}
  \begin{tabular}{lc}
    \toprule
    \textbf{Method} & \textbf{Accuracy (\%)} \\
    \midrule
    SeViLA~\cite{NEURIPS2023_f22a9af8} & 60.9  \\
    IG-VLM~\cite{kim2024image} & 65.3  \\
    VideoTree~\cite{Wang2024VideoTreeAT} & 66.9  \\ 
    LLoVi~\cite{Zhang2023ASL} & 67.1 \\
    TS-LLAVA~\cite{qu2024tsllavaconstructingvisualtokens} & 67.9  \\
    ENTER~\cite{DBLP:journals/corr/abs-2501-14194} & 71.5  \\
    LVNet~\cite{Park2024TooMF} & 71.7  \\
    VideoINSTA~\cite{Liao2024VideoINSTAZL} & 72.8  \\
    \midrule
    \textbf{VideoMultiAgents (Ours)} & \textbf{79.0} \\
    \bottomrule
  \end{tabular}
  \label{tab:intentqa}
\end{table}

\begin{table}
  \caption{Zero-Shot Performance on EgoSchema: A Comparison with State-of-the-Art Approaches.}
  \centering
  \setlength{\tabcolsep}{3pt} 
  \begin{tabular}{p{0.6\columnwidth} c @{~~} c}  
    \toprule
    \multirow{2}{*}{\textbf{Method}} & \multicolumn{2}{c}{\textbf{Accuracy (\%)}} \\ 
    \cmidrule(lr){2-3}
    & \textit{subset} & \textit{fullset} \\
    \midrule
    VideoTree~\cite{Wang2024VideoTreeAT} & 66.2 & 61.1 \\
    LVNet~\cite{Park2024TooMF} & 68.2 & 61.1 \\
    Tarsier~\cite{wang2024tarsierrecipestrainingevaluating} & 68.6 & 61.7 \\
    VideoLLaMA2~\cite{cheng2024videollama2advancingspatialtemporal}  & - & 63.9  \\
    LifelongMemory~\cite{Wang2023LifelongMemoryLL} & 72.0 & 64.7 \\
    LongVU~\cite{shen2024longvuspatiotemporaladaptivecompression} & - & 67.6  \\
    LinVT-Qwen2-VL~\cite{gao2024linvtempowerimagelevellarge} & - & 69.5  \\
    HCQA~\cite{zhang2024hcqaego4degoschema} & 58.8 & \textbf{75.0}  \\
    \midrule
    \textbf{VideoMultiAgents (Ours)} & \textbf{75.4} & 68.0  \\
    \bottomrule
  \end{tabular}
  \label{tab:egoschema}
\end{table}

\section{Ablation Studies}
\label{sec:ablation}

In this section, we conduct ablation studies to analyze the contributions of key components of our proposed method. Specifically, we evaluate the performance of different modalities in the NExT-QA~\cite{xiao2021next} validation set and how different modes of information sharing among agents affect the model's reasoning capability. Additionally, we examine the benefit of Question-Guided Captioning on overall performance.

\begin{figure*}
  \centering
  \includegraphics[width=\textwidth]{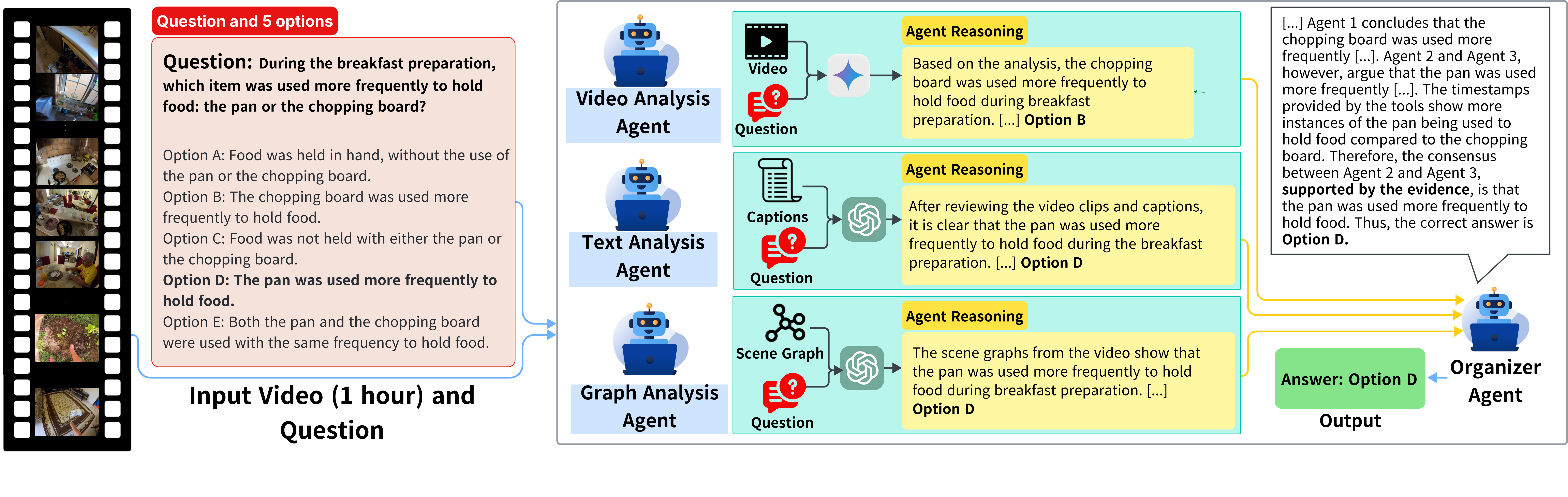}
  \caption{ Inference Process of VideoMultiAgents on Long Videos. This illustration demonstrates our system's analysis of an hour-long video, showcasing the collaborative yet independent reasoning of specialized agents. While the Video Analysis Agent identifies Option B as the answer, both Text and Graph Analysis Agents select Option D with supporting temporal evidence. The Organizer Agent synthesizes these perspectives, examining reasoning quality from each agent before concluding that Option D has stronger evidential support. This example highlights the framework's ability to resolve conflicting interpretations through structured multi-agent collaboration, resulting in accurate understanding even with extended video content.}
  \label{fig:hourvideo}
\end{figure*}

\subsection{Comparison of Single-Agent Modalities}

The upper half of Table \ref{tab:nextqa_val_1000} shows a performance comparison of single-agent approaches across different modalities. 
We observe complementary strengths among the three modalities. Overall, video performs the best, followed by text and graph. However, each modality excels at different categories of questions. Video works the best for descriptive questions (Descriptive Count and Descriptive Other) which focuses on a detailed aspect of the video like counting the number of people in the scene. On the other hand, text excels at causal (Causal How) and temporal questions (Temporal Co-ocurrence, Temporal Present) that require reasoning over many segments of the video while graph excels at descriptive questions related to identification of location (Descriptive Location).

    

\begin{table}
  \caption{Performance Comparison of Different Approaches on 1000 NExT-QA Validation Questions
(CH: Causal-How, CW: Causal-Why, DC: Descriptive-Count, DL: Descriptive-Location, DO: Descriptive-Other, TC: Temporal-Co-ocurrence, TN: Temporal-Previous-Next, TP: Temporal-Present)\\
    {\footnotesize * indicates multi-agent approaches; no asterisk indicates single-agent approaches.}
  }
  \centering
  \setlength{\tabcolsep}{1pt}
  \begin{tabular}{lc|c|c|c|c|c|c|c|c}
    \toprule
    \multirow{2}{*}{\textbf{Approach}} & \multicolumn{9}{c}{\textbf{Accuracy (\%)}} \\ 
    \cmidrule(lr){2-10}
    & \textit{All} & \textit{CH} & \textit{CW} & \textit{DC} & \textit{DL} & \textit{DO} & \textit{TC} & \textit{TN} & \textit{TP} \\
    \midrule
    Text & 77.0 & 85.1 & 78.3 & 73.5 & \textbf{90.6} & 88.7 & 76.4 & 60.2 & \textbf{80.0} \\ 
    Video & 77.6 & 84.3 & 78.6 & \textbf{79.4} & 89.1 & \textbf{94.3} & 73.2 & 65.3 & 60.0 \\ 
    Graph & 72.9 & 76.9 & 76.3 & 52.9 & \textbf{90.6} & 84.9 & 72.4 & 56.3 & 73.3 \\
    \midrule
    Best Category* & 78.5 & 85.1 & 78.6 & \textbf{79.4} & \textbf{90.6} & \textbf{94.3} & 76.4 & 65.3 & \textbf{80.0} \\
    Majority Vote* & 78.7 & 84.3 & 79.6 & 73.5 & \textbf{90.6} & 90.6 & \textbf{78.7} & 65.9 &  73.3 \\
    Star* & 69.5 & 71.6 & 74.6 & 70.6 & 60.9 & 75.5 & 63.8 & 62.5 & 60.0 \\
    Report Star* & 76.5 & \textbf{86.6} & 78.1 & 55.9 & \textbf{90.6} & 92.5 & 70.9 & 64.8 & 60.0 \\
    Debate* & 75.2 & 79.1 & 77.1 & 64.7 & 89.1 & 83.0 & 75.6 & 62.5 & 73.3 \\
    Report* & \textbf{80.4} & 85.1 & \textbf{81.4} & 70.6 & \textbf{90.6} & 90.6 & 78.0 & \textbf{71.6} & \textbf{80.0} \\
    \bottomrule
  \end{tabular}
  \label{tab:nextqa_val_1000}
\end{table}

\subsection{Effectiveness of Multi-Agent Architectures}
\label{subsec:Effectiveness of Multi-Agent Architectures}
To evaluate various multi-agent architectures, we establish two strong, deterministic organizer policies: best category (selecting the modality with best performance per question category) and majority vote (taking votes from all three agents with ties broken by overall performance). As shown in Table \ref{tab:nextqa_val_1000}, majority vote leverages synergies between modalities and results in better performance than choosing the best categories for each question type.

We further examine three primary multi-agent architectures plus a hybrid structure in Table \ref{tab:nextqa_val_1000} and show that the Report architecture outperforms majority vote and other multi-agent architectures. The Report architecture enables the organizer to aggregate independent opinions from all modality agents. Star, on the other hand, grants the organizer more flexibility in initiating conversations with any agent while tracking the entire conversation history. Debate structures the discussion sequentially, with agents able to challenge or support prior claims before the organizer makes a decision.

Our experiments reveal that more structured and independent architectures yield superior performance, with Report performing best, followed by Debate, then Star. To validate this trend, we designed a Report-Star hybrid that combines an initial Report round with a subsequent Star round. The hybrid performed in between Report and Debate, confirming that greater independence improves accuracy.

Analysis of output traces shows that earlier agents tend to bias subsequent conversations, particularly in Star where sequential information flow allows biases to propagate through the organizer and influence later agents. Conversely, Report maintains maximum agent independence, allowing each to form independent opinions that the organizer can then weigh before determining the final answer. This finding aligns with human studies in VQA, where EgoSchema~\cite{mangalam2023egoschema} demonstrates that human annotator accuracy improves by 1.2\% when viewing video without reading questions first compared to unconstrained viewing. Since Star's organizer and Debate's agents can access prior reasoning, they're more susceptible to earlier incorrect answers than human annotators, potentially compounding bias in discussions.


\subsection{Effectiveness of Question-Guided Captioning}

\begin{table}[t]
\centering
\caption{Single-agent performance on NExT-QA using various captioning methods.
We observe improved accuracy with Question-Guided Captioning over LLaVA and GPT-4 baselines.}
\label{tab:caption_comparison}
\begin{tabular}{l c}
\hline
\textbf{Method} & \textbf{Accuracy (\%)} \\
\hline
LLaVA Caption & 68.7 \\
GPT-4o Caption & 74.8 \\
Question-Guided Caption & \textbf{77.0} \\
\hline
\end{tabular}
\end{table}

In this experiment, we evaluate the effectiveness of Question-Guided Captioning, which uses the information from the question in VQA to generate captions. For comparison, we used two image captioning methods: LLaVa~\cite{wang2023videollava}, used in LloVi~\cite{Zhang2023ASL} and GPT-4o~\cite{OpenAI2024GPT4o}, a multimodal model recognized for its strong performance. We conducted our study on the NExT-QA dataset in a single-agent setting that relies solely on caption information, investigating how variations in caption content influence VQA outcomes.

The experimental results are shown in Table~\ref{tab:caption_comparison}. When using the conventional LLaVa-based approach, the score is 68.7\%, while employing GPT-4o yields a score of 74.8\%, confirming GPT-4o’s robust multimodal capabilities as noted in previous research. Moreover, incorporating Question-guided Captioning achieves a score of 77.0\%, surpassing the performance of GPT-4o alone.

Typically, general image captions offer a broad summary of the scene but may not capture the specific details required for VQA, particularly the nouns and verbs found in the question. By incorporating both the question and its answer choices during caption generation, Question-guided Captioning deliberately focuses on the keywords most relevant to the task. As a result, the resulting captions more accurately reflect the information needed for VQA, leading to improved performance overall.

\subsection{Case Study}
\label{sec:casestudy}

We present a case study to showcase the capabilities of VideoMultiAgents for video understanding using the HourVideo~\cite{Chandrasegaran2024HourVideo1V} dataset, which contains hour-long videos. 
In Figure~\ref{fig:hourvideo}, we demonstrate our VideoMultiAgents system analyzing an hour-long video from the HourVideo dataset. The figure shows how each specialized agent independently processes the video and question, then reports its findings to the Organizer Agent for final decision-making. In this particular example, the Video Analysis Agent arrived at an incorrect conclusion due to the limitation of current VLMs in temporal reasoning over event durations. On the other hand, both the Text and Graph Analysis Agents correctly identified the appropriate response because these two modalities explicitly incorporate timestamp information and are thus better at reasoning over durations. The Organizer Agent carefully evaluated the evidence presented by each agent, determining that the timestamps and supporting details provided by the Text and Graph agents were more compelling, ultimately leading to the correct answer. This case study illustrates how our VideoMultiAgents framework effectively handles long-form video content and accurately interprets complex visual information through multi-modal collaboration.

\section{Conclusion}
\label{sec:conclusion}
In this work, we introduced VideoMultiAgents, a multi-agent system in which multiple modality-specialized agents independently perform inference, and an organizer agent integrates their outputs to form the final answer. This structured framework enables comprehensive reasoning across visual, textual, and relational contexts, leading to exceptional effectiveness in understanding long-form videos and handling complex queries. Leveraging complementary advantages of each modality, our system synergizes information from multiple modalities and demonstrate state-of-the-art performance across diverse benchmarks including NExT-QA, Intent-QA and EgoSchema. We believe that this multi-agent paradigm opens up promising directions in video-language understanding, enabling VideoMultiAgents to be readily extended to new tasks and modalities while retaining a coherent and interpretable reasoning process.


{
    \small
    \bibliographystyle{ieeenat_fullname}
    \bibliography{main}

\begin{thebibliography}{36}
\providecommand{\natexlab}[1]{#1}
\providecommand{\url}[1]{\texttt{#1}}
\expandafter\ifx\csname urlstyle\endcsname\relax
  \providecommand{\doi}[1]{doi: #1}\else
  \providecommand{\doi}{doi: \begingroup \urlstyle{rm}\Url}\fi

\bibitem[Ayyubi et~al.(2025)Ayyubi, Liu, Asgarov, Hakim, Sarker, Wang, Tang, Alomari, Atabuzzaman, Lin, Dyava, Chang, and Thomas]{DBLP:journals/corr/abs-2501-14194}
Hammad~A. Ayyubi, Junzhang Liu, Ali Asgarov, Zaber Ibn~Abdul Hakim, Najibul~Haque Sarker, Zhecan Wang, Chia{-}Wei Tang, Hani Alomari, Md. Atabuzzaman, Xudong Lin, Naveen~Reddy Dyava, Shih{-}Fu Chang, and Chris Thomas.
\newblock {ENTER:} event based interpretable reasoning for videoqa.
\newblock \emph{CoRR}, abs/2501.14194, 2025.

\bibitem[Chandrasegaran et~al.(2024)Chandrasegaran, Gupta, Hadzic, Kota, He, Eyzaguirre, Durante, Li, Wu, and Li]{Chandrasegaran2024HourVideo1V}
Keshigeyan Chandrasegaran, Agrim Gupta, Lea~M. Hadzic, Taran Kota, Jimming He, Cristobal Eyzaguirre, Zane Durante, Manling Li, Jiajun Wu, and Fei-Fei Li.
\newblock Hourvideo: 1-hour video-language understanding.
\newblock \emph{ArXiv}, abs/2411.04998, 2024.

\bibitem[Cheng et~al.(2024)Cheng, Leng, Zhang, Xin, Li, Chen, Zhu, Zhang, Luo, Zhao, and Bing]{cheng2024videollama2advancingspatialtemporal}
Zesen Cheng, Sicong Leng, Hang Zhang, Yifei Xin, Xin Li, Guanzheng Chen, Yongxin Zhu, Wenqi Zhang, Ziyang Luo, Deli Zhao, and Lidong Bing.
\newblock Videollama 2: Advancing spatial-temporal modeling and audio understanding in video-llms, 2024.

\bibitem[Cherian et~al.(2021)]{cherian2021adaptive}
Anoop Cherian et~al.
\newblock Adaptive video relationship detection via graph matching and attention.
\newblock In \emph{ICCV}, 2021.

\bibitem[Dai et~al.(2023)Dai, Luo, Durante, Dash, Milstein, Schulman, Adeli, and Fei-Fei]{dai2023momaqa}
Wei Dai, Zelun Luo, Zane Durante, Debadutta Dash, Arnold Milstein, Kevin Schulman, Ehsan Adeli, and Li Fei-Fei.
\newblock Towards fine-grained video question answering.
\newblock \emph{ArXiv arXiv:2303.11331}, 2023.

\bibitem[Fan et~al.(2024)Fan, Ma, Wu, Du, Li, Gao, and Li]{Fan2024VideoAgentAM}
Yue Fan, Xiaojian Ma, Rujie Wu, Yuntao Du, Jiaqi Li, Zhi Gao, and Qing Li.
\newblock Videoagent: A memory-augmented multimodal agent for video understanding.
\newblock \emph{ArXiv}, abs/2403.11481, 2024.

\bibitem[Gao et~al.(2024)Gao, Zhong, Zeng, Tan, Li, and Zhao]{gao2024linvtempowerimagelevellarge}
Lishuai Gao, Yujie Zhong, Yingsen Zeng, Haoxian Tan, Dengjie Li, and Zheng Zhao.
\newblock Linvt: Empower your image-level large language model to understand videos, 2024.

\bibitem[Khan et~al.(2023)Khan, Kuehne, Wu, Chheu, Bousselham, Gan, Lobo, and Shah]{10203761}
Aisha~Urooj Khan, Hilde Kuehne, Bo Wu, Kim Chheu, Walid Bousselham, Chuang Gan, Niels Lobo, and Mubarak Shah.
\newblock Learning situation hyper-graphs for video question answering.
\newblock In \emph{2023 IEEE/CVF Conference on Computer Vision and Pattern Recognition (CVPR)}, pages 14879--14889, 2023.

\bibitem[Kim et~al.(2024)Kim, Choi, Lee, and Rhee]{kim2024image}
Wonkyun Kim, Changin Choi, Wonseok Lee, and Wonjong Rhee.
\newblock An image grid can be worth a video: Zero-shot video question answering using a vlm.
\newblock \emph{IEEE Access}, 2024.

\bibitem[Kugo et~al.(2024)Kugo, Ishibashi, Ono, and Sato]{kugo2024vdma}
Noriyuki Kugo, Tatsuya Ishibashi, Kosuke Ono, and Yuji Sato.
\newblock Vdma: Video question answering with dynamically generated multi-agents.
\newblock \emph{ArXiv}, abs/2407.03610, 2024.

\bibitem[Li et~al.(2023{\natexlab{a}})Li, Li, Savarese, and Hoi]{li2023blip}
Junnan Li, Dongxu Li, Silvio Savarese, and Steven C~H Hoi.
\newblock Blip-2: Bootstrapping language-image pre-training with frozen image encoders and large language models.
\newblock \emph{ICML}, 2023{\natexlab{a}}.

\bibitem[Li et~al.(2023{\natexlab{b}})Li, Wei, Han, and Fan]{10376648}
Jiapeng Li, Ping Wei, Wenjuan Han, and Lifeng Fan.
\newblock Intentqa: Context-aware video intent reasoning.
\newblock In \emph{2023 IEEE/CVF International Conference on Computer Vision (ICCV)}, pages 11929--11940, 2023{\natexlab{b}}.

\bibitem[Liao et~al.(2024)Liao, Erler, Wang, Zhai, Zhang, Ma, and Tresp]{Liao2024VideoINSTAZL}
Ruotong Liao, Max Erler, Huiyu Wang, Guangyao Zhai, Gengyuan Zhang, Yunpu Ma, and Volker Tresp.
\newblock Videoinsta: Zero-shot long video understanding via informative spatial-temporal reasoning with llms.
\newblock In \emph{Conference on Empirical Methods in Natural Language Processing}, 2024.

\bibitem[Lin et~al.(2023{\natexlab{a}})Lin, Zhu, Ye, Ning, Jin, and Yuan]{Lin2023VideoLLaVALU}
Bin Lin, Bin Zhu, Yang Ye, Munan Ning, Peng Jin, and Li Yuan.
\newblock Video-llava: Learning united visual representation by alignment before projection.
\newblock In \emph{Conference on Empirical Methods in Natural Language Processing}, 2023{\natexlab{a}}.

\bibitem[Lin et~al.(2023{\natexlab{b}})Lin, Zhu, Ye, Ning, Jin, and Yuan]{wang2023videollava}
Bin Lin, Bin Zhu, Yang Ye, Munan Ning, Peng Jin, and Li Yuan.
\newblock Video-llava: Learning united visual representation by alignment before projection.
\newblock In \emph{Conference on Empirical Methods in Natural Language Processing}, 2023{\natexlab{b}}.

\bibitem[Lin et~al.(2023{\natexlab{c}})Lin, Ahmed, Li, Lin, Azarnasab, Yang, Wang, Liang, Liu, Lu, Liu, and Wang]{Lin2023MMVIDAV}
Kevin Lin, Faisal Ahmed, Linjie Li, Chung-Ching Lin, Ehsan Azarnasab, Zhengyuan Yang, Jianfeng Wang, Lin Liang, Zicheng Liu, Yumao Lu, Ce Liu, and Lijuan Wang.
\newblock Mm-vid: Advancing video understanding with gpt-4v(ision).
\newblock \emph{ArXiv}, abs/2310.19773, 2023{\natexlab{c}}.

\bibitem[Maaz et~al.(2023)Maaz, Rasheed, Khan, and Khan]{Maaz2023VideoChatGPTTD}
Muhammad Maaz, Hanoona~Abdul Rasheed, Salman~H. Khan, and Fahad~Shahbaz Khan.
\newblock Video-chatgpt: Towards detailed video understanding via large vision and language models.
\newblock In \emph{Annual Meeting of the Association for Computational Linguistics}, 2023.

\bibitem[Mangalam et~al.(2023)Mangalam, Akshulakov, and Malik]{mangalam2023egoschema}
Karttikeya Mangalam, Raiymbek Akshulakov, and Jitendra Malik.
\newblock Egoschema: A diagnostic benchmark for very long-form video language understanding.
\newblock In \emph{Advances in Neural Information Processing Systems (NeurIPS) Datasets and Benchmarks Track}, 2023.

\bibitem[OpenAI(2024)]{OpenAI2024GPT4o}
OpenAI.
\newblock {Hello GPT-4o: OpenAI’s Multimodal GPT-4 Omni Announcement}.
\newblock \url{https://openai.com/hello-gpt-4o}, 2024.
\newblock OpenAI Blog, accessed 13-May-2024.

\bibitem[Park et~al.(2024)Park, Ranasinghe, Kahatapitiya, Ryoo, Kim, and Ryoo]{Park2024TooMF}
Jong~Sung Park, Kanchana Ranasinghe, Kumara Kahatapitiya, Wonjeong Ryoo, Donghyun Kim, and Michael~S. Ryoo.
\newblock Too many frames, not all useful: Efficient strategies for long-form video qa.
\newblock \emph{ArXiv}, abs/2406.09396, 2024.

\bibitem[Qu et~al.(2024)Qu, Li, Tuytelaars, and Moens]{qu2024tsllavaconstructingvisualtokens}
Tingyu Qu, Mingxiao Li, Tinne Tuytelaars, and Marie-Francine Moens.
\newblock Ts-llava: Constructing visual tokens through thumbnail-and-sampling for training-free video large language models, 2024.

\bibitem[Rodin et~al.(2023)Rodin, Furnari, Min, Tripathi, and Farinella]{rodin2023actionscenegraphslongform}
Ivan Rodin, Antonino Furnari, Kyle Min, Subarna Tripathi, and Giovanni~Maria Farinella.
\newblock Action scene graphs for long-form understanding of egocentric videos, 2023.

\bibitem[Shen et~al.(2024)Shen, Xiong, Zhao, Wu, Chen, Zhu, Liu, Xiao, Varadarajan, Bordes, Liu, Xu, Kim, Soran, Krishnamoorthi, Elhoseiny, and Chandra]{shen2024longvuspatiotemporaladaptivecompression}
Xiaoqian Shen, Yunyang Xiong, Changsheng Zhao, Lemeng Wu, Jun Chen, Chenchen Zhu, Zechun Liu, Fanyi Xiao, Balakrishnan Varadarajan, Florian Bordes, Zhuang Liu, Hu Xu, Hyunwoo~J. Kim, Bilge Soran, Raghuraman Krishnamoorthi, Mohamed Elhoseiny, and Vikas Chandra.
\newblock Longvu: Spatiotemporal adaptive compression for long video-language understanding, 2024.

\bibitem[Song et~al.(2023)Song, Chai, Wang, Zhang, Zhou, Wu, Guo, Ye, Lu, Hwang, and Wang]{Song2023MovieChatFD}
Enxin Song, Wenhao Chai, Guanhong Wang, Yucheng Zhang, Haoyang Zhou, Feiyang Wu, Xun Guo, Tianbo Ye, Yang Lu, Jenq-Neng Hwang, and Gaoang Wang.
\newblock Moviechat: From dense token to sparse memory for long video understanding.
\newblock \emph{2024 IEEE/CVF Conference on Computer Vision and Pattern Recognition (CVPR)}, pages 18221--18232, 2023.

\bibitem[Wang et~al.(2024{\natexlab{a}})]{wang2024videoagent}
Chao-hong Wang et~al.
\newblock Videoagent: Long-form video understanding with large language model as agent.
\newblock In \emph{NeurIPS}, 2024{\natexlab{a}}.

\bibitem[Wang et~al.(2024{\natexlab{b}})Wang, Yuan, Zhang, and Sun]{wang2024tarsierrecipestrainingevaluating}
Jiawei Wang, Liping Yuan, Yuchen Zhang, and Haomiao Sun.
\newblock Tarsier: Recipes for training and evaluating large video description models, 2024{\natexlab{b}}.

\bibitem[Wang et~al.(2024{\natexlab{c}})Wang, Zhang, Zohar, and Yeung-Levy]{Wang2024VideoAgentLV}
Xiaohan Wang, Yuhui Zhang, Orr Zohar, and Serena Yeung-Levy.
\newblock Videoagent: Long-form video understanding with large language model as agent.
\newblock \emph{ArXiv}, abs/2403.10517, 2024{\natexlab{c}}.

\bibitem[Wang et~al.(2022)Wang, Li, Li, He, Huang, Zhao, Zhang, Xu, Liu, Wang, Xing, Chen, Pan, Yu, Wang, Wang, and Qiao]{Wang2022InternVideoGV}
Yi Wang, Kunchang Li, Yizhuo Li, Yinan He, Bingkun Huang, Zhiyu Zhao, Hongjie Zhang, Jilan Xu, Yi Liu, Zun Wang, Sen Xing, Guo Chen, Junting Pan, Jiashuo Yu, Yali Wang, Limin Wang, and Yu Qiao.
\newblock Internvideo: General video foundation models via generative and discriminative learning.
\newblock \emph{ArXiv}, abs/2212.03191, 2022.

\bibitem[Wang et~al.(2023)Wang, Yang, and Ren]{Wang2023LifelongMemoryLL}
Ying Wang, Yanlai Yang, and Mengye Ren.
\newblock Lifelongmemory: Leveraging llms for answering queries in long-form egocentric videos.
\newblock In \emph{arXiv}, 2023.

\bibitem[Wang et~al.(2024{\natexlab{d}})Wang, Yu, Stengel-Eskin, Yoon, Cheng, Bertasius, and Bansal]{Wang2024VideoTreeAT}
Ziyang Wang, Shoubin Yu, Elias Stengel-Eskin, Jaehong Yoon, Feng Cheng, Gedas Bertasius, and Mohit Bansal.
\newblock Videotree: Adaptive tree-based video representation for llm reasoning on long videos.
\newblock \emph{ArXiv}, abs/2405.19209, 2024{\natexlab{d}}.

\bibitem[Xiao et~al.(2021)Xiao, Shang, Yao, and Chua]{xiao2021next}
Junbin Xiao, Xindi Shang, Angela Yao, and Tat-Seng Chua.
\newblock {NExT-QA}: Next phase of question-answering to explaining temporal actions.
\newblock In \emph{Proceedings of the IEEE/CVF Conference on Computer Vision and Pattern Recognition (CVPR)}, pages 9777--9786, 2021.

\bibitem[Ye et~al.(2023)Ye, Xu, Xu, Ye, Yan, Zhou, Wang, Hu, Shi, Shi, Li, Xu, Chen, Tian, Qi, Zhang, and Huang]{Ye2023mPLUGOwlME}
Qinghao Ye, Haiyang Xu, Guohai Xu, Jiabo Ye, Ming Yan, Yi Zhou, Junyan Wang, Anwen Hu, Pengcheng Shi, Yaya Shi, Chenliang Li, Yuanhong Xu, Hehong Chen, Junfeng Tian, Qiang Qi, Ji Zhang, and Feiyan Huang.
\newblock mplug-owl: Modularization empowers large language models with multimodality.
\newblock \emph{ArXiv}, abs/2304.14178, 2023.

\bibitem[Yu et~al.(2023)Yu, Cho, Yadav, and Bansal]{NEURIPS2023_f22a9af8}
Shoubin Yu, Jaemin Cho, Prateek Yadav, and Mohit Bansal.
\newblock Self-chained image-language model for video localization and question answering.
\newblock In \emph{Advances in Neural Information Processing Systems}, pages 76749--76771. Curran Associates, Inc., 2023.

\bibitem[Zhang et~al.(2023{\natexlab{a}})Zhang, Lu, Islam, Wang, Yu, Bansal, and Bertasius]{Zhang2023ASL}
Ce Zhang, Taixi Lu, Md~Mohaiminul Islam, Ziyang Wang, Shoubin Yu, Mohit Bansal, and Gedas Bertasius.
\newblock A simple llm framework for long-range video question-answering.
\newblock In \emph{Conference on Empirical Methods in Natural Language Processing}, 2023{\natexlab{a}}.

\bibitem[Zhang et~al.(2023{\natexlab{b}})Zhang, Li, and Bing]{Zhang2023VideoLLaMAAI}
Hang Zhang, Xin Li, and Lidong Bing.
\newblock Video-llama: An instruction-tuned audio-visual language model for video understanding.
\newblock In \emph{Conference on Empirical Methods in Natural Language Processing}, 2023{\natexlab{b}}.

\bibitem[Zhang et~al.(2024)Zhang, Xie, Feng, Li, Liu, and Nie]{zhang2024hcqaego4degoschema}
Haoyu Zhang, Yuquan Xie, Yisen Feng, Zaijing Li, Meng Liu, and Liqiang Nie.
\newblock Hcqa @ ego4d egoschema challenge 2024, 2024.

\end{thebibliography}
}

\end{document}